%% file: main.tex
\documentclass[letterpaper,twocolumn,10pt]{article}
\usepackage{usenix,epsfig,endnotes}
\usepackage[utf8]{inputenc} 
\usepackage[T1]{fontenc}    
\usepackage{cite}
\usepackage{amsmath,amssymb,amsfonts}
\usepackage{algorithmic}
\usepackage{graphicx}
\usepackage{textcomp}
\usepackage{xcolor}
\usepackage{booktabs}
\usepackage{multicol}
\usepackage{hyperref}
\usepackage{subfig}
\usepackage{balance}
\usepackage[title]{appendix}

\def\BibTeX{{\rm B\kern-.05em{\sc i\kern-.025em b}\kern-.08em
    T\kern-.1667em\lower.7ex\hbox{E}\kern-.125emX}}

\newcommand\blfootnote[1]{%
  \begingroup
  \renewcommand\thefootnote{}\footnote{#1}%
  \addtocounter{footnote}{-1}%
  \endgroup
}

\begin{document}

\date{}

\title{\Large \bf Benchmarking Resource Usage for \\ Efficient Distributed Deep Learning}

\author{
{\rm Nathan C. Frey}\thanks{ncfrey@mit.edu}\\
MIT
\and
{\rm Baolin Li}\\
Northeastern University
\and
{\rm Joseph McDonald}\\
MIT
\and
{\rm Dan Zhao}\\
MIT
\and
{\rm Michael Jones}\\
MIT
\and
{\rm David Bestor}\\
MIT
\and
{\rm Devesh Tiwari}\\
Northeastern University
\and
{\rm Vijay Gadepally}\\
MIT
\and
{\rm Siddharth Samsi}\\
MIT
}



\maketitle

\thispagestyle{empty}

\subsection*{Abstract}
Deep learning (DL) workflows demand an ever-increasing budget of compute and energy in order to achieve outsized gains. Neural architecture searches, hyperparameter sweeps, and rapid prototyping consume immense resources that can prevent resource-constrained researchers from experimenting with large models and carry considerable environmental impact.  As such, it becomes essential to understand how different deep neural networks (DNNs) and training leverage increasing compute and energy resources---especially specialized computationally-intensive models across different domains and applications. 

In this paper, we conduct over 3,400 experiments training an array of deep networks representing various domains/tasks---natural language processing, computer vision, and chemistry---on up to 424 graphics processing units (GPUs). During training, our experiments systematically vary compute resource characteristics and energy-saving mechanisms such as power utilization and GPU clock rate limits to capture and illustrate the different trade-offs and scaling behaviors each representative model exhibits under various resource and energy-constrained regimes. We fit power law models that describe how training time scales with available compute resources and energy constraints. We anticipate  that these findings will help inform and guide high-performance computing providers in optimizing resource utilization, by selectively reducing energy consumption for different deep learning tasks/workflows with minimal impact on training. 

\section{Introduction}

As deep learning (DL) workflows become more prevalent in the sciences, neural architecture search and hyperparameter sweeps consume an increasingly enormous amount of compute and power resources~\cite{sharir2020cost, bender2021, zoph2017neural, thompson2020computational, ahmed2020dedemocratization, patterson2021carbon, patel2020} at high-performance computing (HPC) centers~\cite{Strubell_Ganesh_McCallum_2020, yin2019, adolf2016} and cloud providers. While the cost per training step has decreased for deep neural networks (DNNs) due to optimized hardware and backend optimizations, overall costs have increased and training large models can reach into the millions of dollars~\cite{sharir2020cost}. Traditionally, HPC centers limit GPU usage to prevent users from misusing systems, while cloud providers eagerly allow users to provision as many resources as they can afford. Rarely do scientific DL practitioners examine their resource needs; most workflows are either run on a single GPU due to the lack of engineering infrastructure needed to scale, or are run on the maximum number of available GPUs~\cite{samsi2021mit, jeon2019}. Efficient training and scaling strategies may be even more important than architecture details in some domains~\cite{bello2021revisiting, tan2020efficientnet, radosavovic2020designing}. To complicate matters, given the fundamentally different nature of their respective tasks, in order to better allocate limited compute resources, various domains (e.g., NLP, vision) each have their own preferred model architectures and optimization strategies, likely resulting in what is very different scaling behavior across different models/tasks. To better allocate limited compute resources, HPC centers and users need a simple way to estimate the scaling behavior of their models and identify the best, most scalable model implementation for their application and the optimal amount of compute to provision. 

\blfootnote{DISTRIBUTION STATEMENT A. Approved for public release. Distribution is unlimited. This material is based upon work supported by the Under Secretary of Defense for Research and Engineering under Air Force Contract No. FA8702-15-D-0001. Any opinions, findings, conclusions or recommendations expressed in this material are those of the author(s) and do not necessarily reflect the views of the Under Secretary of Defense for Research and Engineering. © 2021 Massachusetts Institute of Technology. Delivered to the U.S. Government with Unlimited Rights, as defined in DFARS Part 252.227-7013 or 7014 (Feb 2014). Notwithstanding any copyright notice, U.S. Government rights in this work are defined by DFARS 252.227-7013 or DFARS 252.227-7014 as detailed above. Use of this work other than as specifically authorized by the U.S. Government may violate any copyrights that exist in this work.}

\begin{figure*}[t]
    \centering
    \includegraphics[scale=0.6]{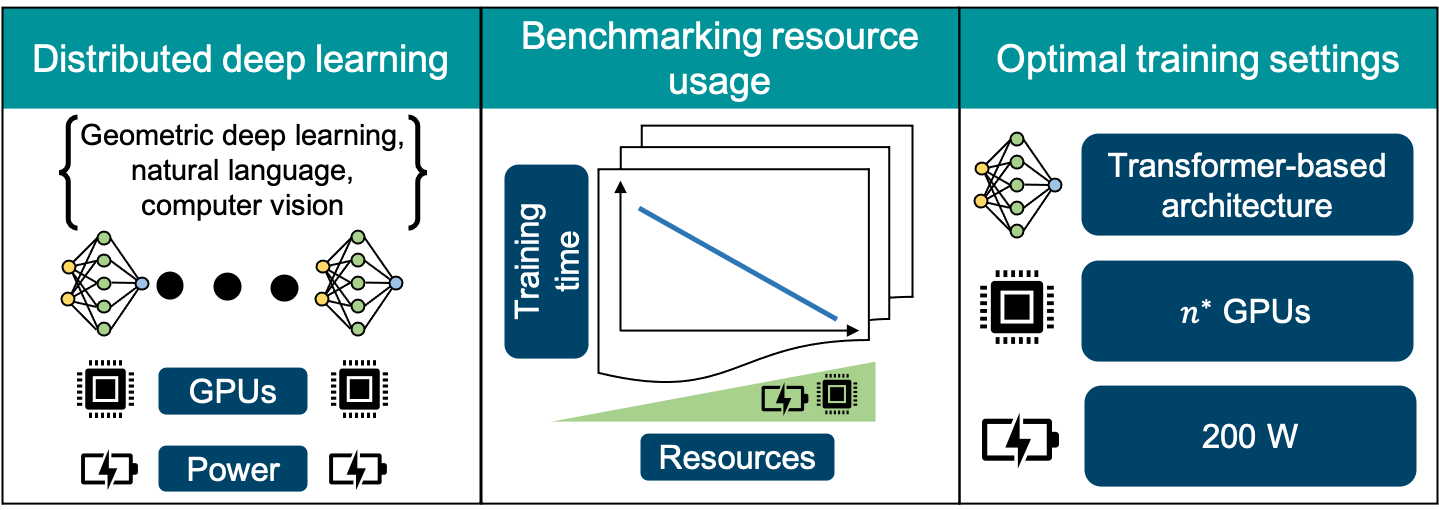}
    \caption{\textbf{Benchmarking experiments training DNNs on more than 400 GPUs with controlled power consumption reveal optimal settings for efficient distributed DL.} Over 3,400 distributed training experiments show that transformer-based models and graph neural networks with directional message passing exhibit superior utilization of increased computational resources, while restricting GPU power consumption to 200 W reduces total energy consumption without slowing down training. All models see diminishing returns from distributed training at high GPU counts due to communication bottlenecks.}
    \label{fig:overview}
\end{figure*}

The complexity of DNNs and the variety of numerical libraries and hardware accelerators ~\cite{reuther2020} available make predicting the execution time of training a model challenging. Previous efforts  estimated training times with linear models depending on the number of floating point operations per epoch ~\cite{Qi2017}, while others have leveraged DNNs themselves to learn the non-linear relationship ~\cite{justus2018} between network architecture, the data manifold, computational infrastructure, and execution time. More recent work to predict the execution time of fine-tuning DNNs uses a linearized approximation of the dynamics of a DNN during fine-tuning ~\cite{zancato2020}, following the Neural Tangent Kernel (NTK) approach ~\cite{jacot2018}. These methods may yield impressive accuracy in training time estimation, at least in terms of number of training steps required, but they are cumbersome and impractical for daily usage in an HPC center. More relevantly, these approaches do not account for energy consumption, which is difficult to estimate for general network configurations ~\cite{GARCIAMARTIN201975, yang2017method}, or variation in GPU utilization. Scientists with limited HPC experience and rapidly changing DL workflows need guidance from large-scale, distributed DL training experiments to optimize resource allocation for efficient, scalable deep learning. 

In this paper, we train six different, representative DL models (Table \ref{model_table}) with applications across computer vision (CV), natural language processing (NLP), and geometric deep learning (GDL) and investigate their scaling behavior across hundreds of GPUs. We monitor GPU utilization and energy consumption during distributed training and identify optimal settings for efficient training and opportunities for improved scaling and hardware utilization. Our main goal is not to generate precise predictions of execution time, but instead to study the impacts of and the relationship between model architecture and compute utilization on distributed training time. By comparing model architectures via their scaling exponents, we can estimate training times for variations on common architectures such as convolutional neural networks (CNNs), transformer-based language models, and graph neural networks (GNNs). This will help scientific DL practitioners in developing methods to better profile different model architectures and determine the most time and energy-efficient workflow for their own hardware configurations. 

To the best of our knowledge, current literature on scaling experiments for DL has not focused on the effects of energy-consumption strategies such as power limiting the hardware or changing clock frequencies of the GPU to limit performance. We hope that these findings will also help enable predictions of model-scaling behavior on performance-limited hardware to potentially anticipate the energy needs for different classes of DNNs in future work. 

\section{Methods and Experimental Setup}

\paragraph{Environment} All experiments described in this paper were conducted on an operational, petascale supercomputing system. The cluster consists of 448 compute nodes with dual Intel Xeon Gold 6248 CPUs with 384 GB of RAM and two NVIDIA Volta V100 GPUs with 32 GB of memory per node. A graphical summary of the experiments and insights presented in this paper is shown in Figure \ref{fig:overview}. All models were trained using GPUs.

\paragraph{Model Selection} Our choices of benchmarks and models were informed by our operational experiences from the workloads and usage patterns characteristic of the system---we chose DNN models and training tasks that are representative of common user workloads, but also models that tend to be frequent workhorses for research, testing, and experiments in the broader research community. We also sought to capture a diverse range of "classic" and state-of-the-art DL architectures that are widely used across different domains. Our intent in this work is not to propose a new benchmark or evaluate the relative merit of one benchmark against another. Rather, our goal is to investigate and illustrate the scaling behavior of representative model architectures, based on computational workloads on an operational supercomputing system, and develop simple descriptions of training execution time. The specific models in our experiments are listed in Table~\ref{model_table} and can be grouped into three domains or categories characterized by the commonality of their tasks, similarity of model architectures, or both. 

\begin{table}[htbp]
\caption{Deep Neural Network Models}
\begin{center}
\begin{tabular}{cll}
\toprule
\textbf{Domain/Model Class}&\textbf{Model}&\textbf{Reference} \\
\midrule
 & VGG16&\cite{vgg16}\\
Computer Vision (CV) & Inceptionv3&\cite{inceptionv3}\\
 & ResNet50&\cite{he2016deep}\\
\midrule
Language (NLP) & BERT&\cite{devlin2019bert}\\
\midrule
Geometric & DimeNet&\cite{klicpera2020directional}\\
&SchNet&\cite{schutt2017schnet}\\
\bottomrule
\end{tabular}
\label{model_table}
\end{center}
\end{table}

\paragraph{Experimental Design and Metrics}

Throughout our experiments, for each model in Table~\ref{model_table}, we collect time-series data on their GPU memory utilization and streaming multiprocessor (SM) utilization, training speed, and total energy consumption/expenditure throughout each of their training runs under different GPU power caps (100 W, 200 W, and 250 W), GPU clock rates (135 MHz, 735 MHz, and 1380 MHz), and number of GPUs (2, 4, 8, 16, 32, 64, etc.) aggregated on a per-epoch level. When a power or clock rate cap is applied, it is applied uniformly and simultaneously to all GPUs used in training. Model training was performed on NVIDIA Tesla V100 GPUs and GPU utilization; power consumption etc., were monitored using the \texttt{nvidia-smi} tool. Training speed or training time is calculated as wall-clock time per epoch, measured in seconds, and power/energy consumption (or training energy)  measured in joules, which is calculated by multiplying wall-clock time spent in training by the power consumption measured in watts.

We detail the specific training settings and configurations for each model by domain/model class in the sub-sections below. We note that while different training techniques and optimized hyperparameter selections can both shorten training time and improve model performance, our goal is neither to achieve the best possible performance on benchmarks nor to train until we achieve a pre-specified level of performance. Given the diverse array of DL models we train and their representative domains, their evaluative criteria vary as well, measuring fundamentally different model qualities on non-equivalent scales. For instance, while top-$k$ accuracy and Intersection over Union (IoU) are used for image classification and object detection respectively, perplexity and bits-per-character (BPC) are used to gauge model performance in NLP tasks; 
as such, we leave these as potential avenues for future work. 

Instead, the training settings for our different representative models are motivated by our aim to: (1) collect sufficiently large amounts of utilization and energy data from each domain/model class during training for meaningful, wide-coverage analysis, as well as to (2) document and illustrate trends and trade-offs between training time, compute resources, and energy constraints reflective of an average user training scenario where users may lack the knowledge, ability, or resources to specify the best training settings for reaching a desired level of performance as quickly as possible, or where users are running exploratory experiments and exercises. Examples of these types of situations include transfer-learning experiments, evaluation of new dataset benchmarks, rapid prototyping and evaluation of new model architectures, situations where researchers collaborate/experiment with models across various domains, and more.

All of our training datasets and model implementations are publicly available. Similarly, our DL workload data from these experiments will be made publicly available for the broader research community.

\subsection{Geometric Deep Learning (GDL)}
Unlike traditional learning where the focus is on tabular, sequential, or image data, GDL focuses on models that specifically operate on non-Euclidean structures such as graphs and manifolds ~\cite{bronstein2017, bronstein2021geometric}. As an example, GNNs are widely used in chemistry and materials science ~\cite{Kearnes_2016} where condensed phases of matter are represented as graphs of atoms (nodes) and chemical bonds (edges). 

For simplicity and consistency, we used two open-sourced, state-of-the-art GNN architectures available in PyTorch Geometric for molecular machine learning applications ~\cite{Fey/Lenssen/2019}: DimeNet (2.1M parameters) ~\cite{klicpera2020directional} and SchNet (455K parameters) ~\cite{schutt2017schnet}. The DimeNet and SchNet model hyperparameters are taken as default from PyTorch Geometric. DimeNet uses an embedding dimension of 128, 6 interaction blocks, 7 spherical harmonics, and 6 radial basis functions. SchNet uses an embedding dimension of 128, 128 convolutional filters, 6 interaction blocks, and 50 gaussians. All GNNs were trained using the Distributed Data Parallel (DDP) accelerator ~\cite{li2020pytorch} and PyTorch Lightning ~\cite{falcon2019pytorch} for multi-GPU training using the \emph{LitMatter} framework ~\cite{frey2021scalable}. For our data, we use the QM9 dataset for small molecules ~\cite{ramakrishnan2014quantum} during training. We train our GNNs for 200 epochs with a fixed batch size of 128 per GPU across all power capping and clock speed settings/experiments. For SchNet, we train for 1,000 epochs at only the maximum power and clock speed setting in order to collect sufficient utilization data for analysis on those settings. Both DimeNet and SchNet were trained using a learning rate of $10^{-3}$ with Adam ~\cite{kingma2017adam}. 

\subsection{Natural Language Processing (NLP)}

Language modeling is a set of approaches for obtaining distributions over sequences of words and is an important first step towards many common NLP tasks. Models are typically ``pre-trained" using self-supervised learning methods for predicting word sequences before applying them to supervised tasks such as entailment or question answering.
While much attention is given to the most accurate language models, they can require considerably long training times and significant energy usage ~\cite{sharir2020cost, bender2021}.

For our language model representative, we trained Google's Bidirectional Encoder Representations from Transformers (BERT) language model \cite{devlin2019bert} with masked language modeling, which involves predicting randomly chosen word tokens that are obscured from a large training set of text. We use a popular PyTorch implementation\footnote{\url{https://github.com/huggingface/transformers/blob/master/examples/pytorch/language-modeling/run\_mlm.py}} for masked language modeling from Hugging Face and the WikiText-103 dataset for training \cite{wikitext}.
BERT-Base language models (110M parameters) were trained on a range of GPUs for 4, 6, 10, 15, 25, or 40 epochs, depending on the number of GPUs to collect adequate time series data on GPU utilization for each experiment. A per-GPU batch size of 8 was used for all BERT experiments. BERT was trained with the AdamW optimizer and a learning rate of $5\times10^{-5}$.
This implementation uses PyTorch's DistributedDataParallel modules with Message Passing Interface (MPI) facilitating inter-node communication. Due to the large size of BERT-Base, the computational cost of training, and the importance of BERT within the NLP domain, we  consider BERT as a representative transformer-based language model.

\subsection{Computer Vision}
In much work on image recognition and vision-related tasks, convolutional kernels have been key to early successes, given the domain of data (i.e., images) that lends naturally to convolutional priors. These DL models for computer vision and image classification continue to be widely used, either in classification tasks or as backbone networks in applications such as object detection, semantic segmentation, and many more. In this paper, we use open source implementations of popular models for image classification: VGG16 (138M parameters) ~\cite{vgg16}, ResNet50 (25M parameters) ~\cite{he2016deep}, and Inceptionv3 (23M parameters) ~\cite{inceptionv3}, available from \cite{tfbenchmarks}. All three models were trained with a data-parallel training approach leveraging the Horovod framework \cite{horovod}. The training data used was the ImageNet \cite{imagenet} dataset, converted into TFRecord format using utilities provided by the benchmark implementation described above. Models were trained in mixed-precision format for 10 epochs, over a range of GPUs distributed across multiple nodes, with a per-GPU batch size of 256, an initial learning rate of $10^{-5}$, and stochastic gradient descent as the optimizer. For all models, the number of epochs was chosen to ensure that adequate utilization data could be collected for each experiment in the low-GPU and high-GPU limits.

\input{./char.tex}

\section{Fitting Scaling Relationships}
\begin{figure}[htbp]
\centerline{\includegraphics[width=0.48\textwidth]{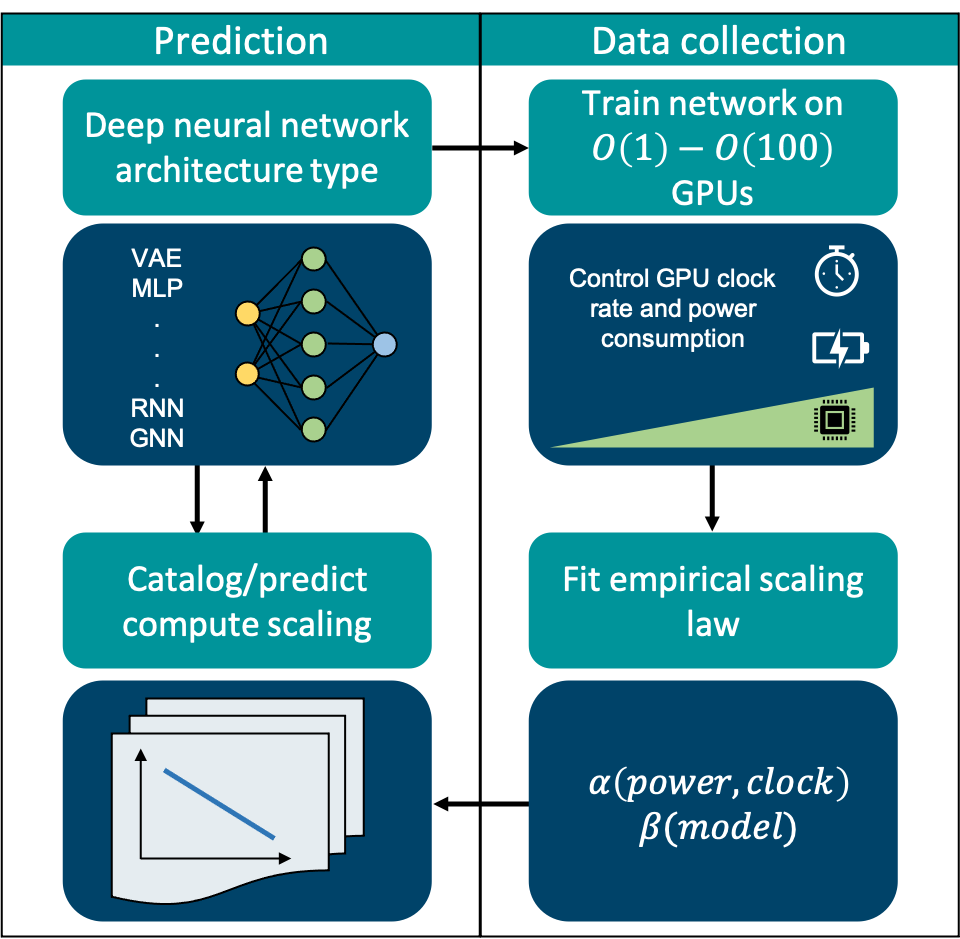}}
\caption{\textbf{Workflow for determining scaling relationships that describe training time dependence on number of GPUs and GPU characteristics}. Empirical scaling laws, if present and sufficiently robust/extrapolative, can help inform anticipatory scaling strategies and energy-performance trade-offs for managing and monitoring distributed training.}
\label{workflow}
\end{figure}

\paragraph{Motivation} In this section, we fit and derive statistically significant, robust power laws that describe how distributed training time scales with the number of GPUs utilized in training using the data collected from our training experiments in Section 3. The workflow is presented diagrammatically in Figure \ref{workflow}; we note that this framework is easily extendable to experiments with other models beyond those in our paper.

The purpose of our model is to describe relationships that exist in these training time and energy consumption trade-offs, not to precisely predict job execution time nor predict training completion times.

 
\paragraph{Model Fitting and Specification} To test and assess our hypothesized power-law relationship between the number of GPUs and training time across our experiments, we fit an empirical model that models training time as a function of several explanatory variables via a power-law functional form. More specifically, with $t$ as training time per epoch, $m$ representing the model architecture, $p$ the maximum allowable amount of power drawn from the GPU(s), $c$ the maximum allowable GPU clock rate, $N$ as the number of GPU(s) utilized, and $\varepsilon$ as an all-inclusive term accounting for all other training and data-dependent characteristics (e.g., hyper-parameter and other training settings), our model follows the form:
\begin{equation} \label{eq:powerlaw}
t \approx f(m, p, c, N, \varepsilon) = \alpha(\cdot) \hspace{0.5mm} N^{-\beta(\cdot)}
\end{equation}
where $\alpha(\cdot)$ and $\beta(\cdot)$ are a multiplicative and exponential factor, respectively, that govern the strength of the scaling relationship between $t$ and $N$. In particular, we focus on $\beta$, which we refer to as the characteristic \textit{scaling exponent}, as it describes the strength or degree to which training time scales with the number of GPUs. In other words, on a log-log scale, $\alpha$ is an additive component independent of $N$, while $\beta$ is the slope or the multiplicative component of the effect of $N$ on $t$---our main object of interest. 
As such, when in log-log scale, changes in $\alpha$ will induce parallel shifts in the curve representing the relationship between training time and GPU count (i.e., across all GPU counts) and changes in $\beta$ will affect the curvature/slope of said curve (i.e., rate of change between train time and GPU count). Note $\alpha$ and $\beta$ can also be functions of some of the explanatory variables described. By varying $p$ and $c$ in our experiments, we can simulate and study the effects of strategies that implicitly vary $\alpha$ and $\beta$ that accelerate (or decelerate) training time per epoch. The training time per epoch data versus number of GPUs is then fit to a power law (Equation \ref{eq:powerlaw}), which describes the training time over several orders of magnitude of compute. 

Given our primary focus on the scaling relationship and trade-offs between training time and number of GPUs across different power and clock settings, we make $N$ and $\beta$ the focal points of our functional power-law form. We re-emphasize that our aim is an intentionally simplified model that focuses on the scaling relationships across different DL models and their respective domains/model classes, and leave other extensions for future work.


To fit the models, we run log-log regressions by taking the natural log of both sides of Equation \ref{eq:powerlaw} and estimating the parameters via least squares, using training time per epoch as our response variable given our explanatory variable ($N$, number of GPUs). Scikit-learn ~\cite{pedregosa2011scikit} was used to perform the model fitting and \emph{statsmodels} ~\cite{seabold2010statsmodels} was used for the statistical analyses. We fit said model for each neural network architecture in each domain/model class across each power cap and clock rate setting we performed from our experiments. To measure the goodness of fit, we calculate the coefficient of determination, $R^2$.


The simplified expression in Equation \ref{eq:powerlaw} allows us to narrow our focus but, as such, it does not explicitly capture important variation in hyper-parameters like batch size and model size which may affect training time scaling. 
We also attempted log-normal fits and confirmed that the power law specification better describes the relationships in the data, but more data and maximum likelihood fitting may lead to different parameter estimates ~\cite{clauset2009}. We catalog the scaling relations for each of our representative models and detail our findings and results per model class/domain below. \textit{Overall, the high $R^2$, the large absolute magnitude of our estimated scaling exponents $\beta$, and the statistical significance of $\beta$ from our fitted log-log regressions lend confidence to the notion that we are capturing real, meaningful scaling phenomena rather than statistical noise.}


\begin{table*}[htbp]
    \centering
    \caption{Log-log models regressing training time per epoch of each model against number of GPUs at 250 W power and 1380 MHz clock rate (i.e., without any power/clock caps). $\beta_{\text{\# GPUs}}$ indicates the scaling exponent and each column corresponds to a log-log regression fit of a model's training time per epoch for each number of GPUs used across its training.}
    \begin{tabular}{ccccccccc}
    \toprule
    & \multicolumn{2}{c}{Geometric Deep} & &  \multicolumn{1}{c}{Natural Language } & &  \multicolumn{3}{c}{Computer Vision} \\
    & \multicolumn{2}{c}{Learning} & &  \multicolumn{1}{c}{Processing} & &  \multicolumn{3}{c}{} \\
    \midrule
    \textbf{Model} &  DimeNet & SchNet & & BERT & & ResNet50 & VGG16 & InceptionV3 \\
    \cline{2-3} \cline{5-5} \cline{7-9}
    $\beta_{\text{\# GPUs}}$     & 0.82 & 0.42 && 0.87 && 0.52 & 0.64 & 0.44 \\
    \textbf{Goodness-of-fit $R^2$} & 0.99 & 0.90 && 0.97 && 0.95 & 0.98 & 0.93 \\
    \bottomrule \\
    \end{tabular}
    \label{tab:combined_table1}
\end{table*}

\subsection{Geometric Deep Learning}
\begin{figure}[htbp]
\includegraphics[width=0.45\textwidth]{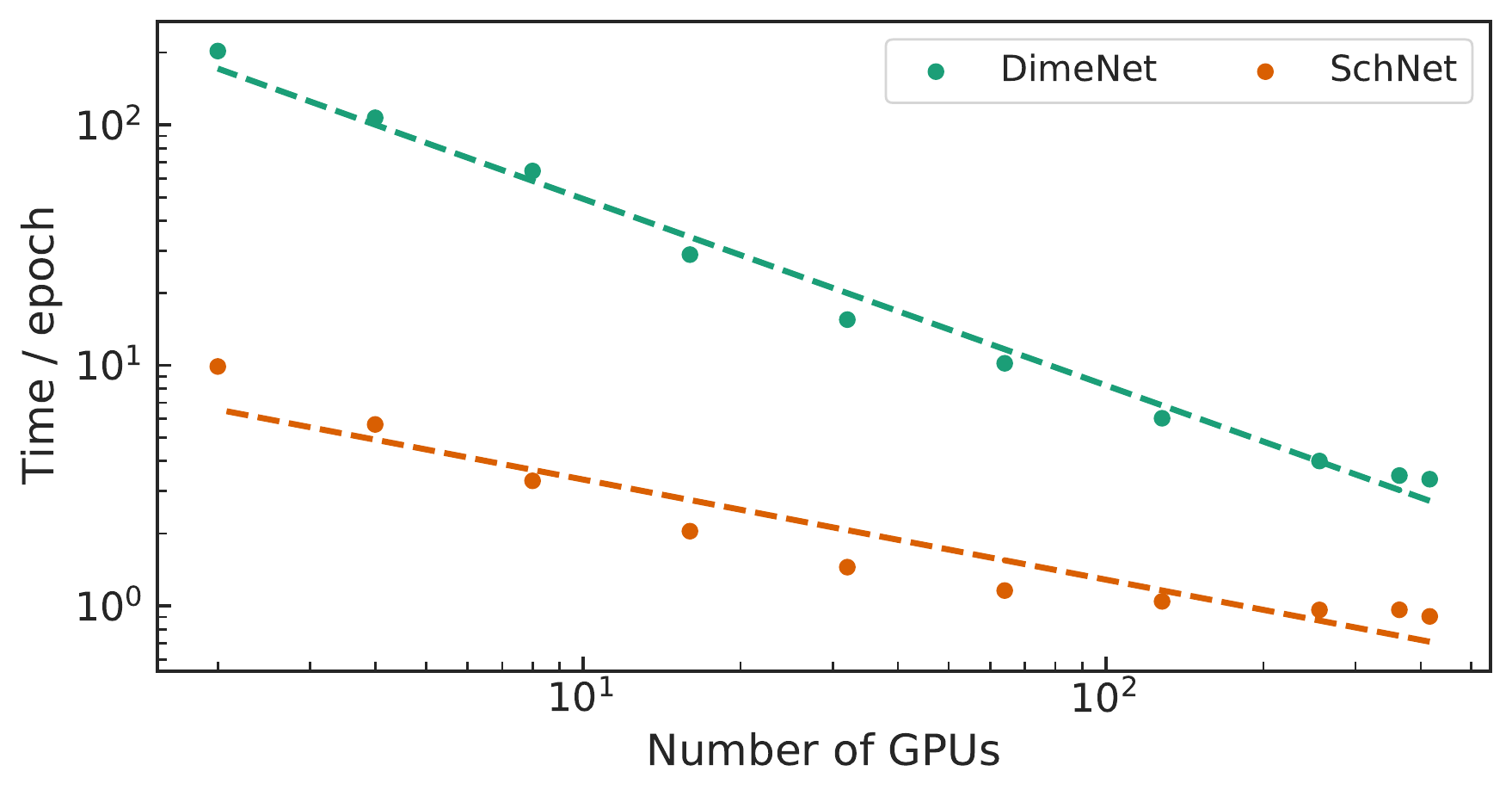}
\caption{(GNNs) \textbf{GNNs with directional message passing exhibit significantly larger speedups with distributed training.} Training time (in seconds) scaling for GNNs. Note the steeper slope of DimeNet (in green).}
\label{gnn_scaling}
\end{figure}

\noindent\newline\textbf{Significant architectural differences can change scaling, even within classes of network topologies. }To illustrate the sensitivity of compute scaling to model topology, even within domains/model classes, we show the scaling behavior for our two GNN architectures in Figure \ref{gnn_scaling}: DimeNet ~\cite{klicpera2020directional} and SchNet ~\cite{schutt2017schnet}. Both operate on molecular graph data and are trained on the Quantum Machine 9  (QM9) dataset ~\cite{ramakrishnan2014quantum} but DimeNet is a GNN that incorporates rotationally equivariant directional message passing, while SchNet uses continuous-filter convolutions---results are shown in Table \ref{tab:combined_table1}. 

Both models exhibit power law scaling up to 416 GPUs with an $R^2 \geq 0.90$. DimeNet's scaling exponent is $0.82 \pm 0.03$ and SchNet's scaling exponent $0.42 \pm 0.05$. The 0.40 difference in the scaling exponents and the low standard error indicate that the difference in scaling between the models is significant. From the diagram in Figure \ref{gnn_scaling} and results in Table \ref{tab:combined_table1}, we clearly see that DimeNet, with additional parameters and rotational equivariance, exhibits a much greater speedup (60$\times$) per epoch as we scale up the number of GPUs.

\begin{figure}[htbp]
\includegraphics[width=0.48\textwidth]{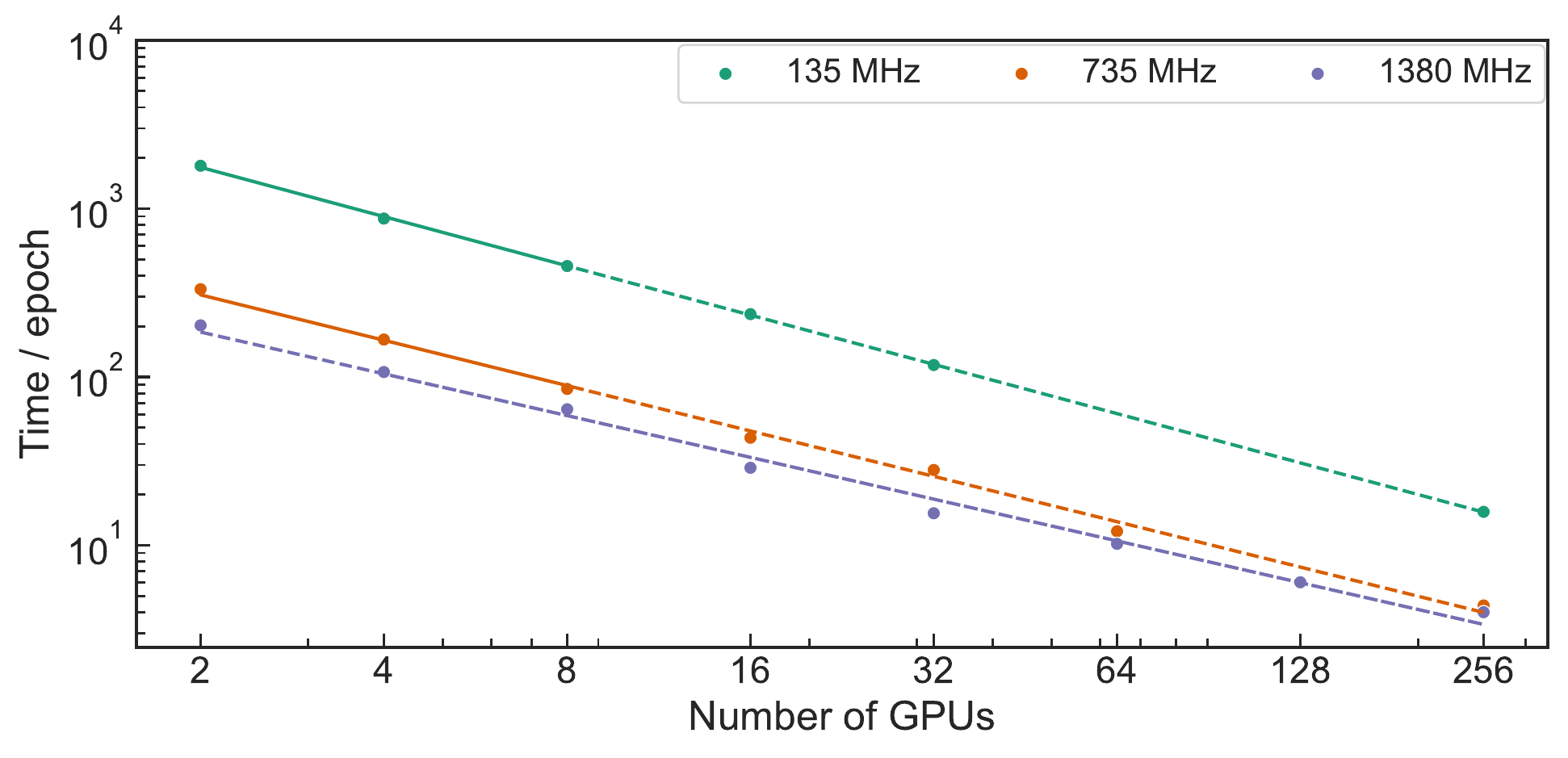}
\caption{(GNNs, DimeNet) \textbf{Reducing the GPU clock rate increases the offset (non-distributed training time), but does not affect the slope of distributed training time scaling.} Training time scaling for DimeNet at different GPU clock rates.}
\label{dimenet_clock_scaling}
\end{figure}

\noindent\newline\textbf{Power and clock speed caps do not significantly impact scaling exponents.} We vary clock speed $c$ between three setting caps (135, 735, and 1380 MHz) to simulate different levels of GPU utilization for DimeNet (Figure \ref{dimenet_clock_scaling}) and SchNet, trained on up to 256 GPUs. We find that limiting the clock rate has a small effect on the scaling exponent $\beta$ for DimeNet (Table \ref{dimenet_clock_table}), with higher values of $c$ resulting in smaller values of $\beta$. As expected, decreasing $c$ increases the multiplicative constant $\alpha$, reflecting slower training times for reduced GPU utilization. Therefore, reducing the clock rate \emph{shifts} the training time scaling curve down ($\alpha$), but does not alter the slope of the curve ($\beta$), demonstrating that \textit{the effects of changing/capping clock rates do not depend on the number of GPUs}. Due to communication bottlenecks, training times for SchNet clearly deviate from power law scaling beyond 64 GPUs for all measured clock rates. Under 64 GPUs, the same behavior as seen for DimeNet is observed, with clock rate having a minimal impact on training time scaling.

\begin{table}[htbp]
\caption{DimeNet log-log regressions of training time per epoch on number of GPUs under different clock speed limits but max power (250 W).}
\begin{center}
\begin{tabular}{cccc}
\toprule
\textbf{Clock speed (MHz)}&\textbf{$\beta_{\text{\# GPUs}}$}&\textbf{Goodness-of-fit $R^2$} \\ 
\midrule
135 & 0.97 & 1.0 \\
735 & 0.90 & 0.99 \\
1380 & 0.82 & 0.99 \\
\bottomrule
\end{tabular}
\label{dimenet_clock_table}
\end{center}
\end{table}

\begin{figure}[htbp]
\includegraphics[width=0.48\textwidth]{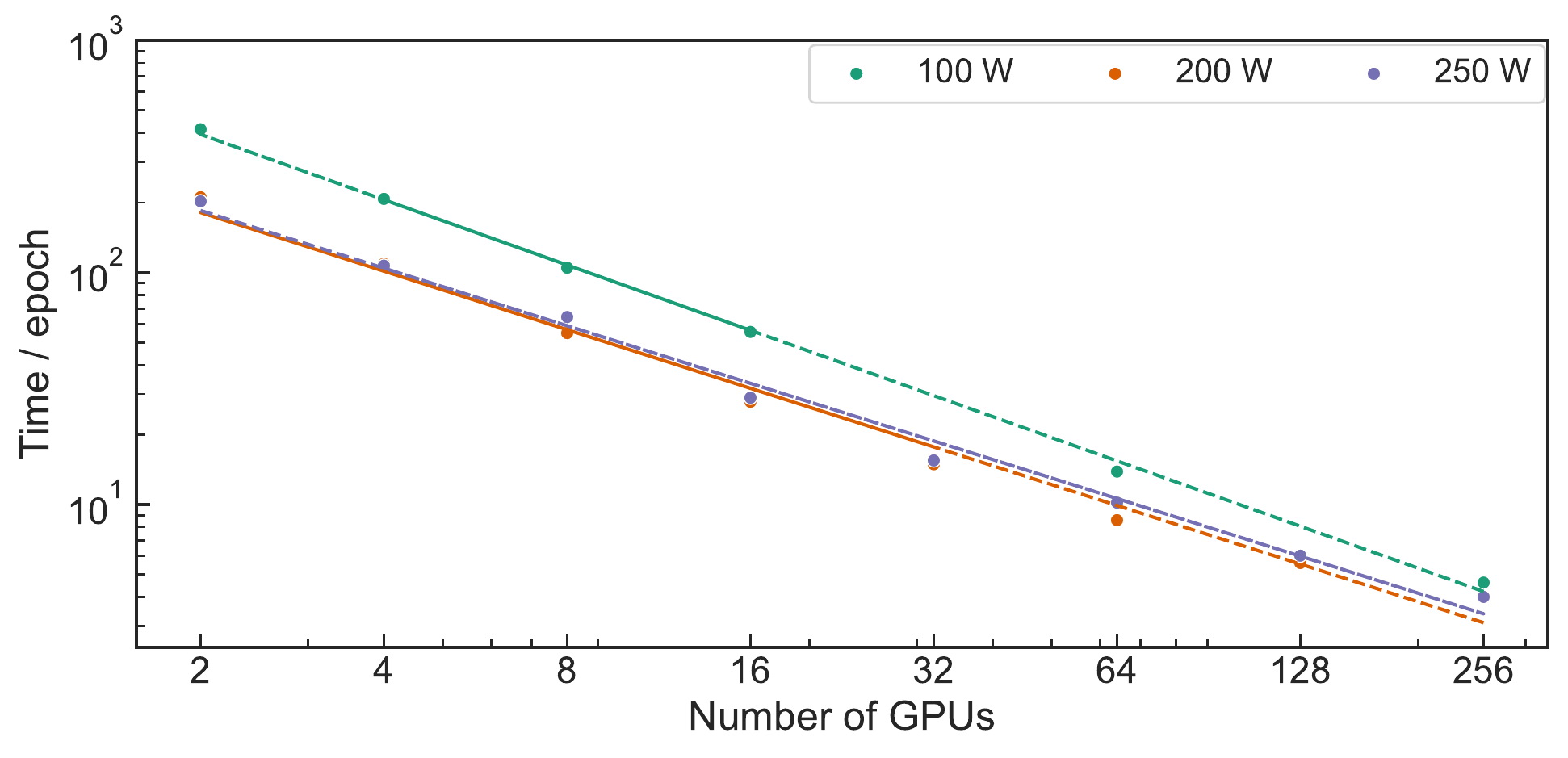}
\caption{(GNNs, DimeNet) \textbf{Reducing the GPU power below 200 W increases the offset (non-distributed training time), but does not affect the slope of distributed training time scaling.} Training time scaling for DimeNet at different GPU power caps.}
\label{dimenet_power_scaling}
\end{figure}

Scaling training up to 256 GPUs at various power caps (100, 200, and 250 W) are shown for DimeNet in Figure \ref{dimenet_power_scaling}. For DimeNet the power cap has a minimal impact on the scaling exponent $\beta$ (Table \ref{dimenet_power_table}), and training times are nearly identical for 200 and 250 W. An increase in training times is seen at 100 W, reflecting under-utilization of GPUs and a larger $\alpha$ parameter. SchNet demonstrates near-invariance to different levels of power capping, due to communication bottlenecks and consistent under-utilization of GPU resources. We note that although power cap and GPU clock rate induce similar effects, we systematically investigate varying both mechanisms for completeness in order to determine optimal settings for energy savings.

\begin{table}[htbp]
\caption{DimeNet log-log regressions of training time per epoch on number of GPUs under different power caps but max clock speed (1380 MHz).}
\begin{center}
\begin{tabular}{cccc}
\toprule
\textbf{Power (W)}&\textbf{$\beta_{\text{\# GPUs}}$}&\textbf{Goodness-of-fit $R^2$} 
\\
\midrule
100 & 0.93 & 0.99 \\ 
200 & 0.84 & 0.99 \\ 
250 & 0.82 & 0.99 \\ 
\bottomrule
\end{tabular}
\label{dimenet_power_table}
\end{center}
\end{table}



\subsection{Natural Language Processing}
\begin{figure}[htbp]
\includegraphics[width=0.45\textwidth]{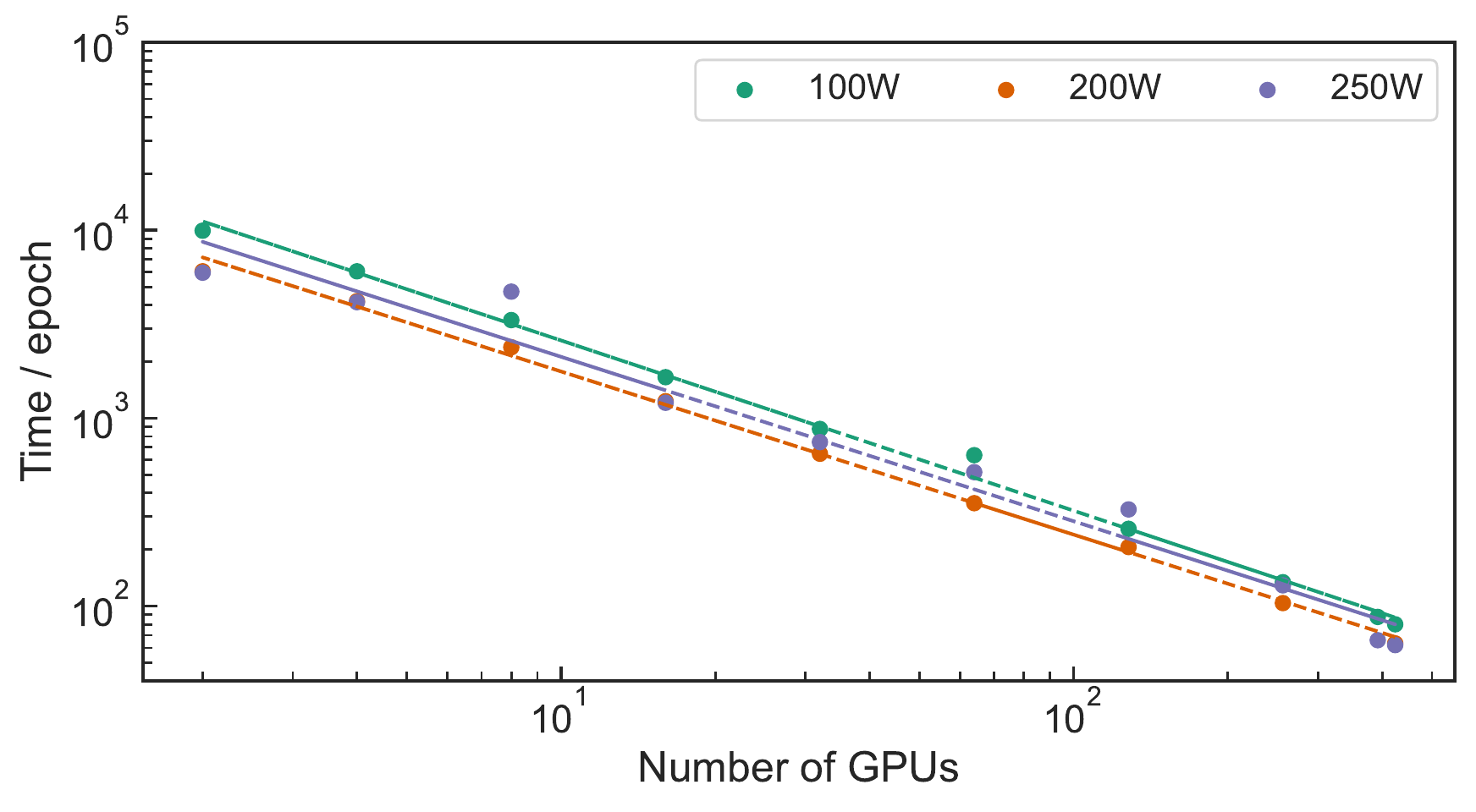}
\caption{\textbf{BERT exhibits optimal distributed training time scaling, training time is minimally impacted by GPU power restrictions.} Training time scaling for BERT NLP models at different GPU power caps.}
\label{nlp_scaling}
\end{figure}

\noindent\newline\textbf{Transformer-based models show optimal compute scaling. }We investigated the impact of GPU power utilization on compute scaling when training a BERT language model
with 110M parameters (BERT-Base) on the WikiText-103 dataset.
Figure \ref{nlp_scaling} shows the compute scaling of BERT training at three different values of $p$ using between 2 and 424 GPUs, where $p \in \{100W, 200W, 250W\}$. At any value of $p$, all GPUs are capped at $p$ as the maximum power setting and similarly for clock rate limits $c$. These power caps simulate lower-power hardware and models that are unable to fully utilize the maximum power of a GPU, and also have implications for the energy efficiency of model training. 

From the scaling exponents in Table \ref{nlp_table}, we show that the power cap has a minimal impact on compute scaling, over a wide range of $p$ values, just as observed for the graph neural networks. Instead, $p$ is more impactful on $\alpha$, influencing shifts in train time per epoch in a way that is invariant to, or across all, GPU counts. Interestingly, allowing more power consumption (i.e., increasing $p$) does not always translate into shorter training time; for BERT, restricting the power usage to 200 W (50 W below the maximum possible power usage) leads to optimal training times below 424 GPUs. We observed training time speedups of up to 76$\times$ through multi-GPU training. At maximum power (250 W), BERT also exhibits the best scaling of any model considered here. The scaling exponent $\beta$ for BERT is $0.87 \pm 0.03$ (Table \ref{tab:combined_table1}), while the next best scaling network is DimeNet with $\beta = 0.82 \pm 0.03$.

\begin{table}[htbp]
\caption{BERT log-log regressions of training time per epoch on number of GPUs under different power caps.}
\begin{center}
\begin{tabular}{cccc}
\toprule
\textbf{Power cap (W)}&\textbf{$\beta_{\text{\# GPUs}}$}&\textbf{Goodness-of-fit $R^2$}& \\ 
\midrule
100 & 0.91 & 0.99 \\ 
200 & 0.86 & 0.99 \\ 
250 & 0.87 & 0.99 \\ 
\bottomrule
\end{tabular}
\label{nlp_table}
\end{center}
\end{table}

\subsection{Computer Vision}
\begin{figure}[htbp]
\includegraphics[width=0.45\textwidth]{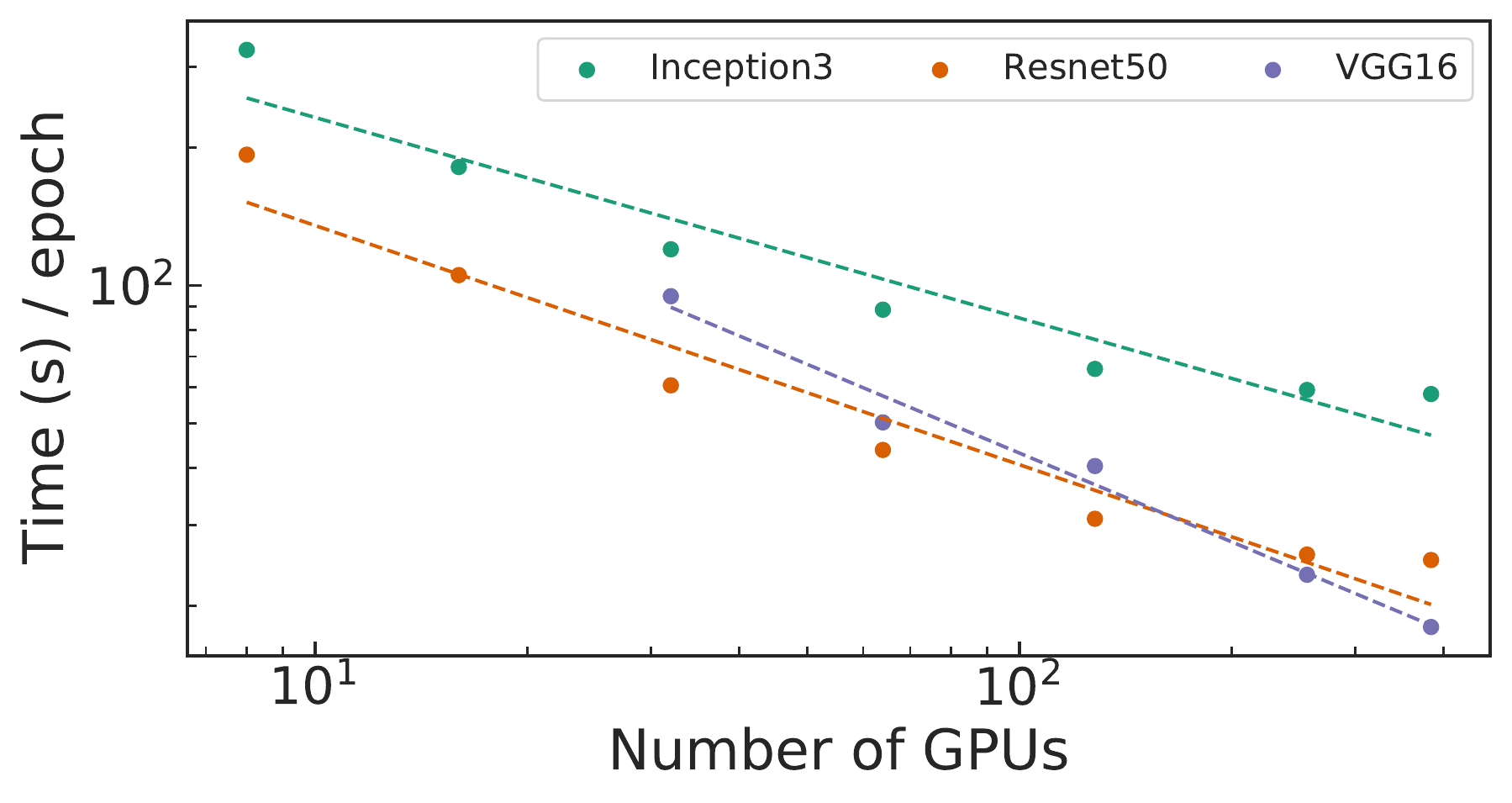}
\caption{\textbf{ResNet50 demonstrates superior training time scaling below 100 GPUs, but VGG16 has better scaling performance beyond 100 GPUs.} Training time scaling for vision.}
\label{vision_scaling}
\end{figure}

\noindent\newline\textbf{Convolutional neural networks exhibit similar compute scaling behavior.} The compute scaling results in Figure \ref{vision_scaling} show the minimal differences between CNN models, despite the differences in implementations, size, and architectural details. Across these different architectures, the maximum difference in scaling exponent is 0.20. VGG16 shows the best scaling with increasing compute, although each architecture demonstrates an overall training time speedup of 5$\times$ to 6$\times$ through multi-GPU training. 

\begin{table}[htbp]
\caption{ResNet50 log-log regressions of training time per epoch on number of GPUs under different power caps.}
\begin{center}
\begin{tabular}{cccc}
\toprule
\textbf{Power cap (W)}&\textbf{$\beta_{\text{\# GPUs}}$}&\textbf{Goodness-of-fit $R^2$}& \\ 
\midrule
100 & 0.83 & 1.0 \\ 
200 & 0.83 & 0.99 \\ 
250 & 0.84 & 0.99 \\ 
\bottomrule
\end{tabular}
\label{vision_table}
\end{center}
\end{table}

\begin{figure}[!t]
\includegraphics[width=0.45\textwidth]{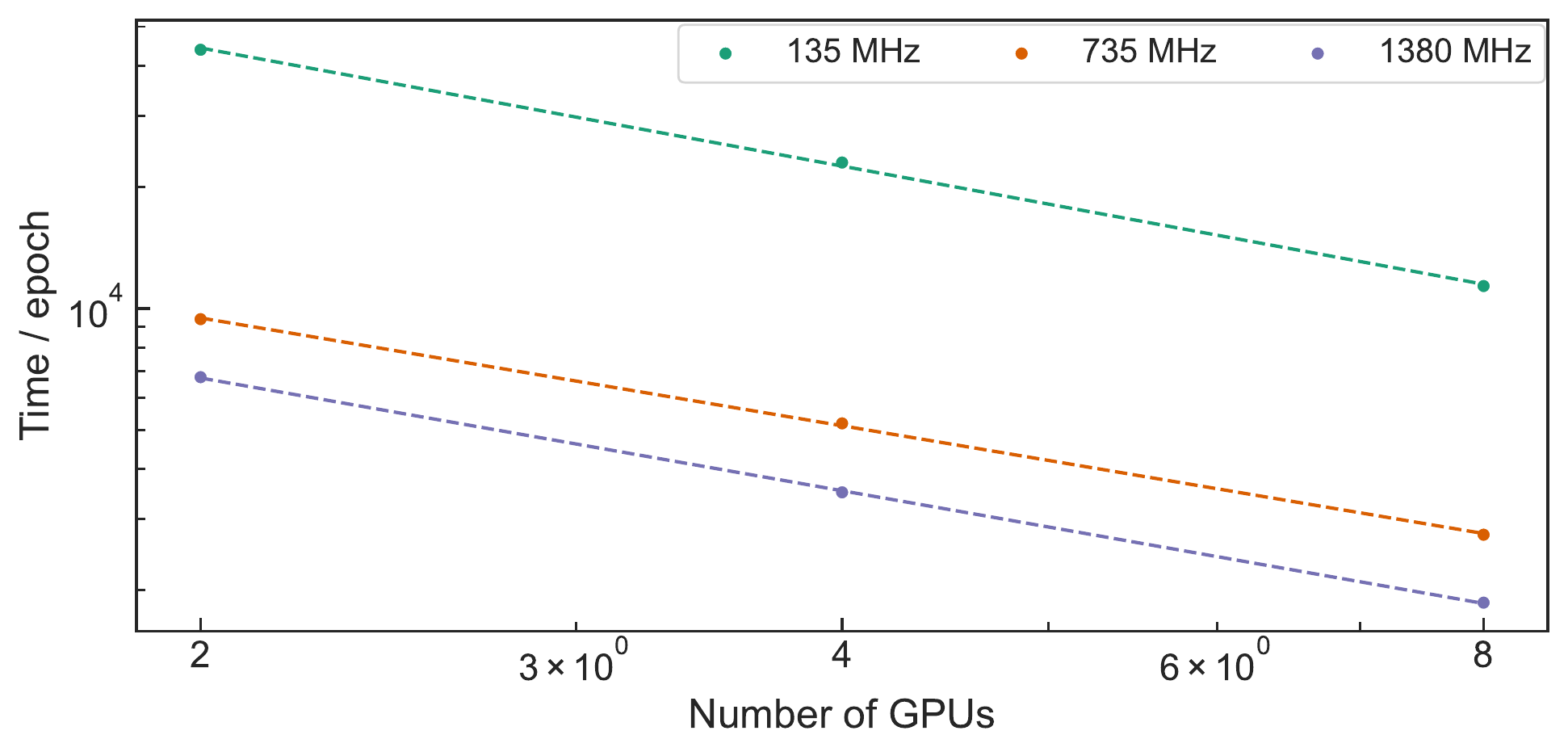}
\caption{\textbf{Reducing the GPU clock rate increases the offset (non-distributed training time), but does not affect the slope of distributed training time scaling.} Training time scaling for ResNet50 at different GPU clock rates.}
\label{resnet_clock_scaling}
\end{figure}

\begin{figure}[!t]
\includegraphics[width=0.45\textwidth]{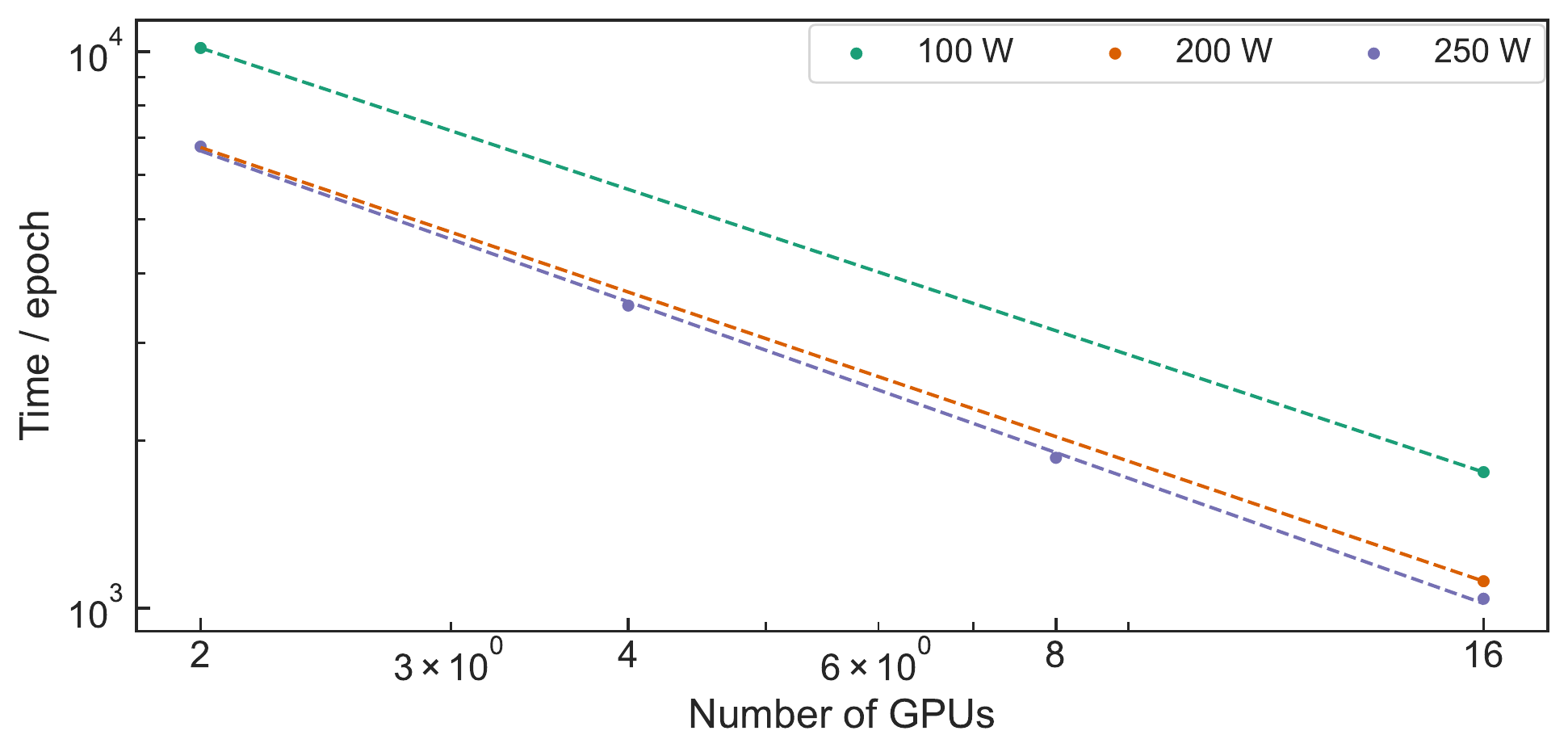}
\caption{\textbf{Reducing the GPU power below 200 W increases the offset (non-distributed training time), but does not affect the slope of distributed training time scaling.} Training time scaling for ResNet50 at different GPU power caps.}
\label{resnet_power_scaling}
\end{figure}

The same shift observed for GNN and BERT models in $\alpha$, while $\beta$ values remain stable, is seen for CNN architectures when changing the clock speed limit from 135 MHz to 1380 MHz and changing the power cap from 100 to 250 W (Table \ref{vision_table}). Representative plots for clock speed and power capping experiments are shown for ResNet50 in Figures \ref{resnet_clock_scaling} and \ref{resnet_power_scaling}.


\section{Discussion} \label{discussion}
We conducted and reported results from over 3,400 experiments training six different, representative DNN architectures (DimeNet, SchNet, BERT, VGG16, Inceptionv3, and ResNet50) on up to 424 GPUs, monitoring and analyzing GPU utilization and energy usage under a variety of GPU power and clock rate settings. We then use our experimental data to derive simple power-law scaling relations that describe how training time per epoch changes as a function of compute resource availability (i.e., number of GPUs) and different settings in power consumption and clock rates.

A general conclusion for all deep networks considered is that reducing the GPU power cap leads to significant energy savings with minimal adverse impacts on training speed/time. Our experiments show that 200 W is an optimal power cap for geometric, language, and vision models that decreases total energy consumption, on average, by at least 10\% with a negligible decrease in training speed (i.e., negligible increase in training time per epoch). The results of this work are being used in an operational, peta-scale supercomputing system to optimally allocate compute resources to minimize training time for DL models, maximize utilization of compute, and decrease energy usage.

We also find that all models under-utilize GPU memory and display significant variation in memory utilization during distributed training, which may be indicative of an opportunity for software and model architecture optimizations to fully utilize modern hardware when scaling up GPUs for training.

Finally, our empirical model, fit on our experimental data, shows that training time scaling can be well-described by power laws which are robust to training time behavior up to hundreds of GPUs across many of our representative deep networks. We show that our empirical power-law model can serve as a preliminary diagnostic tool to simulate and evaluate how different energy-efficient mechanisms affect these scaling behaviors, which we hope can be improved upon in future work; for instance, through this framework, we demonstrate that reducing the GPU clock rate and power consumption has a negligible influence on the scaling of distributed training, but has a significant impact on the overall training time execution. Lastly, our framework offers a way to quantitatively capture and compare scaling behaviors of how well different networks leverage increasing resources via their respective scaling exponents. 

We anticipate that our findings will inform HPC centers, cloud providers, researchers, and future work looking to develop tools to optimize resource allocation with energy-savings in mind. For example, HPC centers may increase compute allocation to users who incorporate the energy saving settings and scalable network architectures described in this work into their workflows. Our future work will investigate the full carbon footprint of these workloads in further detail, along with the impacts of optimized scaling on neural network performance (i.e., training to a desired level of performance). Given the intensive compute resources required to conduct such scaling studies, we intend to make all experimental data from this study publicly available. 
as part of the MIT Supercloud Datacenter Challenge ~\cite{samsi2021mit} via \href{https://dcc.mit.edu/}{\textcolor{blue}{this https URL}}.

\section*{Acknowledgment}
The authors acknowledge the MIT SuperCloud ~\cite{reuther2018interactive} and Lincoln Laboratory Supercomputing Center for providing HPC and consultation resources that have contributed to the research results reported within this paper.
The authors acknowledge the MIT SuperCloud team: William Arcand, David Bestor, William Bergeron, Chansup Byun, Matthew Hubbell, Michael Houle, Mike Jones, Jeremy Kepner, Anna Klein, Peter Michaleas, Lauren Milechin, Julie Mullen, Andrew Prout, Albert Reuther, Antonio Rosa, and Charles Yee. The authors also wish to acknowledge the following individuals for their contributions and support: Bob Bond, Allan Vanterpool, Tucker Hamilton, Jeff Gottschalk, Tim Kraska, Mike Kanaan, Charles Leiserson, Dave Martinez, John Radovan, Steve Rejto, Daniela Rus, Marc Zissman.

\section*{Availability}
The dataset from all deep learning workloads presented in this paper is publicly available via the web at \href{https://dcc.mit.edu/}{\textcolor{blue}{this https URL}}. 

\balance

{\footnotesize \bibliographystyle{acm}
\bibliography{bib}}


\newpage
\begin{appendices}

\section{Supplementary Figures} \label{appendix:figs}

Due to computational resource constraints, GPU utilization and energy consumption data at 200 and 100 W power caps are shown below only for BERT and ResNet50.

\begin{figure}[!htb]
    \centering
    \includegraphics[scale=0.48]{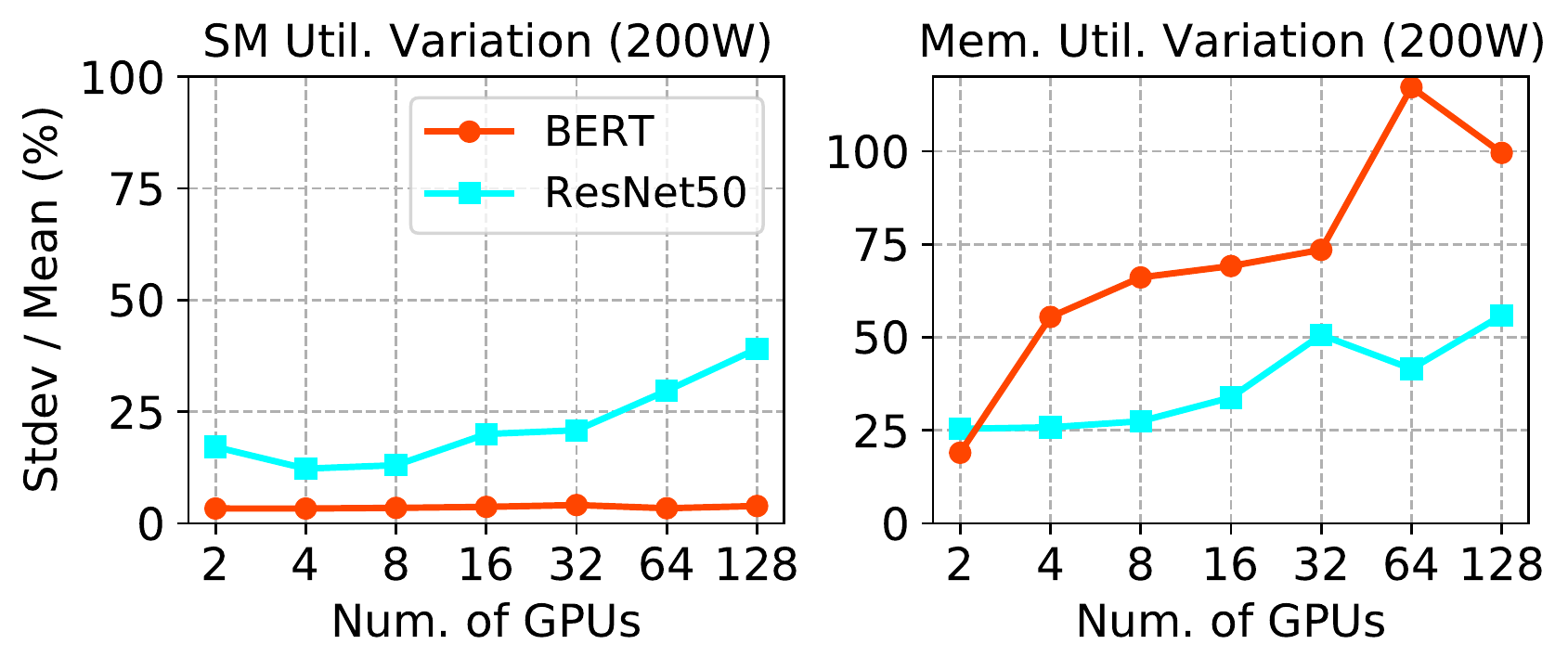}
    \caption{\textbf{Variation in GPU utilization during training increases with an increasing number of GPUs at a 200 W power cap.}}
    \label{fig:char3_200}
\end{figure}

\begin{figure}[!htb]
    \centering
    \includegraphics[scale=0.48]{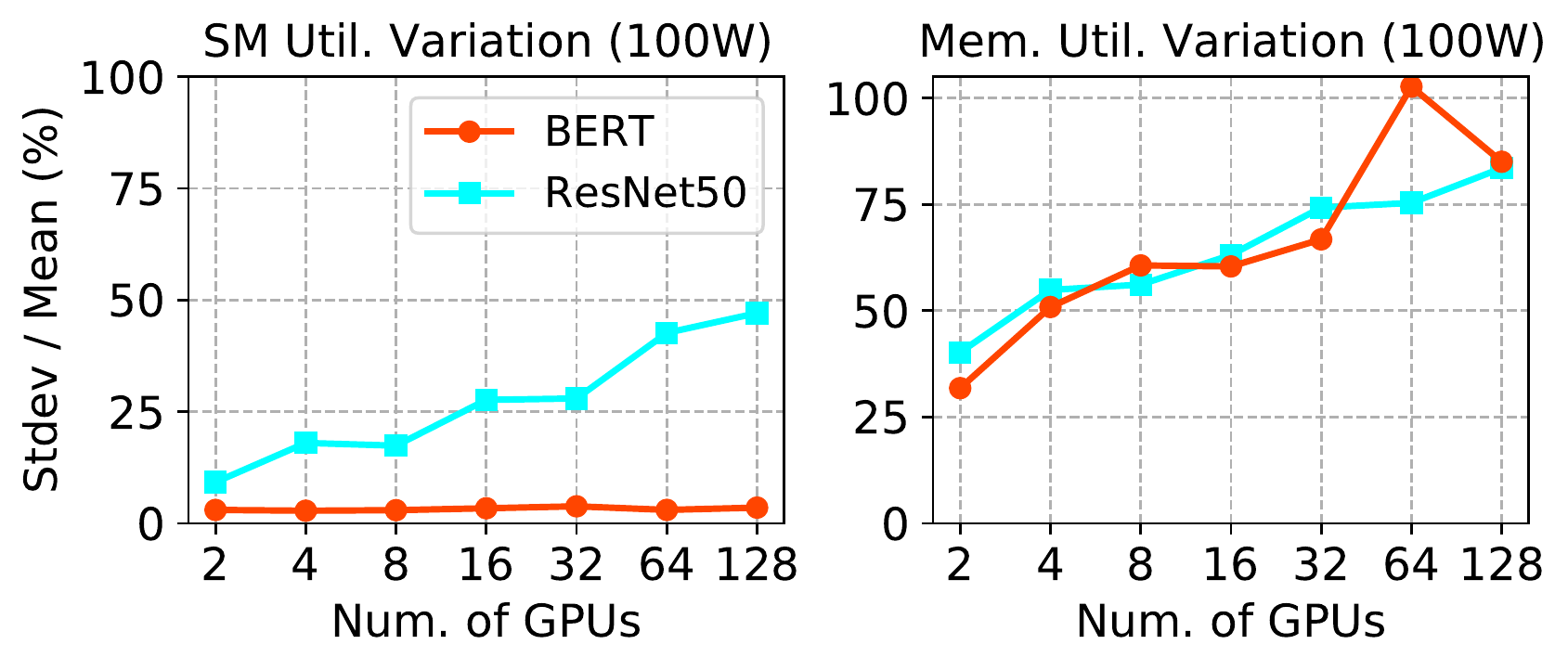}
    \caption{\textbf{Variation in GPU utilization during training increases with an increasing number of GPUs at a 100 W power cap.}}
    \label{fig:char3_100}
\end{figure}

\begin{figure}[!htb]
    \centering
    \includegraphics[scale=0.44]{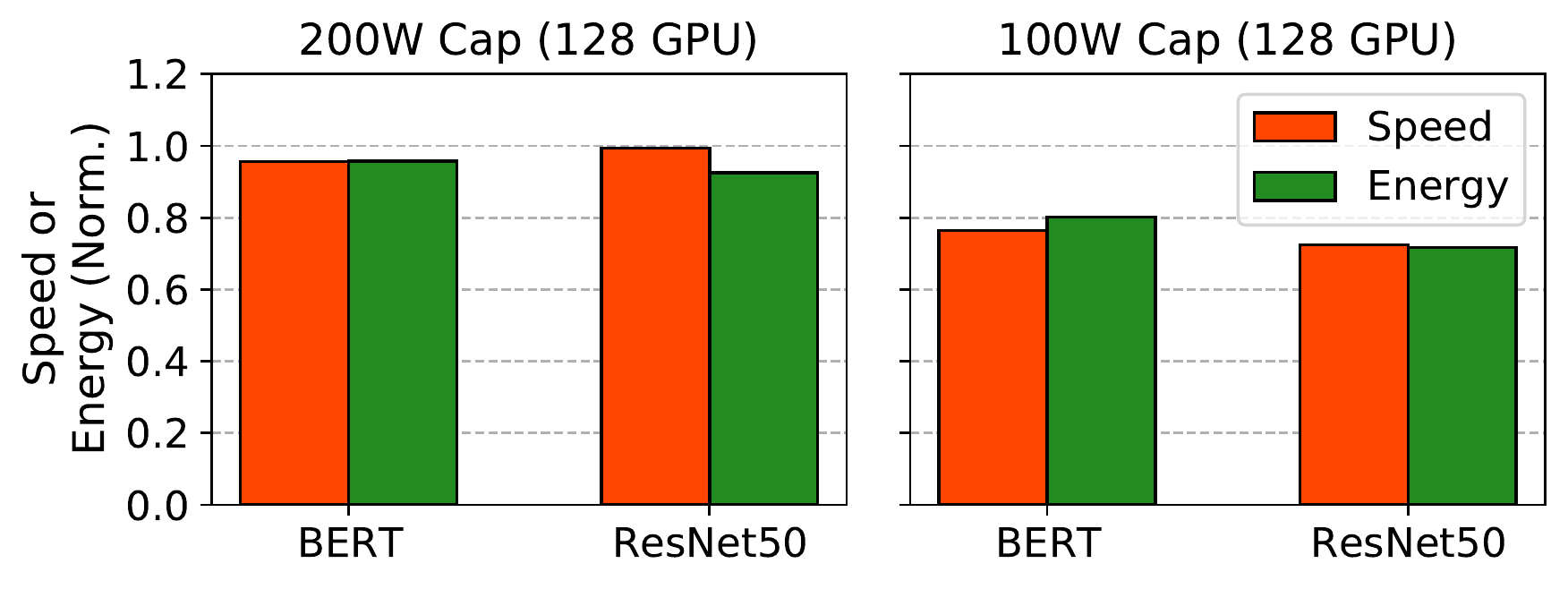}
    \caption{\textbf{The effects of power capping to 200 W (reduced energy consumption without impacting training speed) persist with distributed training at 128 GPUs.} Training speed and total training energy on different GPU power capping settings when trained with 128 GPUs. Values are normalized to training speed and energy without power capping (250 W).}
    \label{fig:char5_128gpu}
\end{figure}

\end{appendices}

\end{document}

%% file: char.tex
\section{Experimental Results and Findings} \label{experiments}

Below, we first describe and discuss some initial findings from our distributed training experiments for each model and its representative domain/model class. 

\begin{figure}[t]
    \centering
    \includegraphics[scale=0.42]{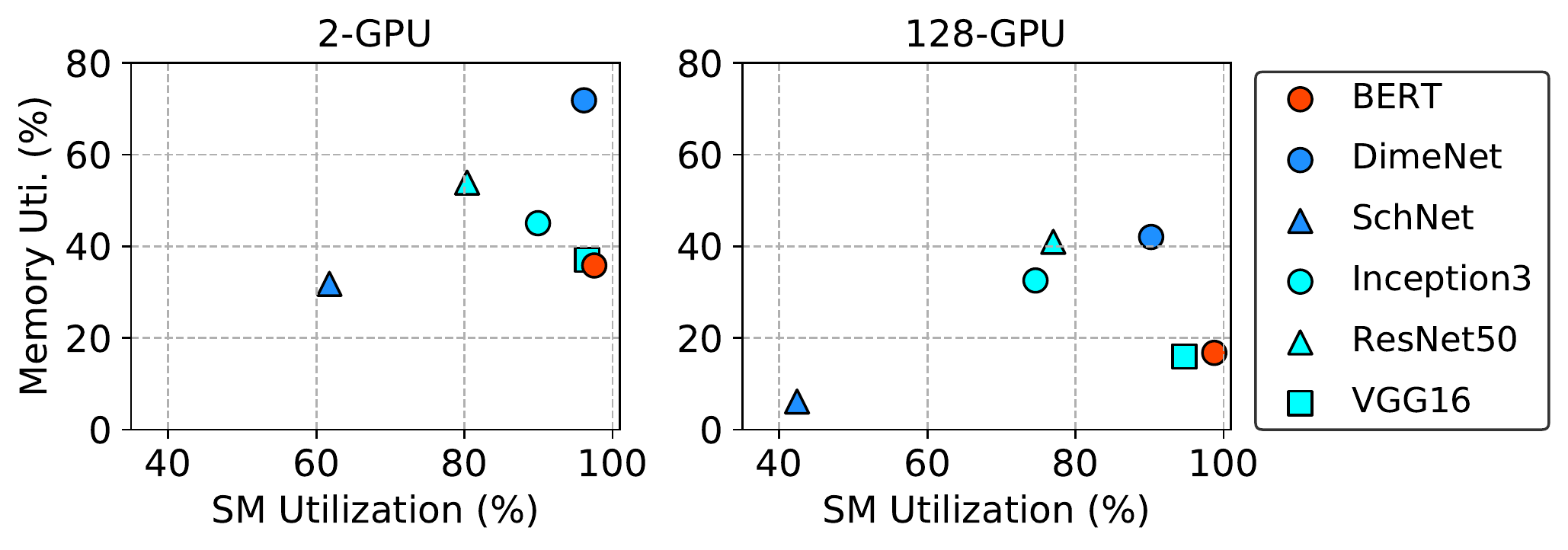}
    \caption{\textbf{All models show reduced memory utilization with increasing number of GPUs, and GNNs show reduced streaming multiprocessor (SM) utilization.} Joint GPU memory and SM utilization characteristics for NLP, GNN, and vision models during training, changes when scaling with different number of GPUs (2 GPUs vs. 128 GPUs). }
    \label{fig:char1}
\end{figure}

\noindent\newline\textbf{When scaling up the number of GPUs, changes in GPU-utilization are heavily architecture-dependent, even within domains/model classes; however, all models show reduced memory utilization during training}. Naturally, different models are likely to exhibit different GPU utilization profiles and these differences are likely to become more pronounced as differences in network architecture, and therefore computational pathways, become more pronounced. In Figure ~\ref{fig:char1}, we show the joint GPU memory and SM utilization characteristics for the models described in Table~\ref{model_table} at two representative scales of distributed training (on 2 GPUs versus 128 GPUs). Geometric models are shown in blue, BERT is shown in red, and vision models are shown in teal, where memory and SM utilization are averaged over all GPU workers.

Although GPU utilization varies both across and within model classes/domains, we note some common behaviors and trends when we scale up training. When the models are trained on 2 GPUs in parallel on a single node, nearly all models tend to have high SM (>60\%) and memory utilization (>30\%). As training is scaled up to 128 GPUs across 64 nodes, all models shift to lower memory utilization; however, we see that the GNNs also see lower SM utilization alongside memory utilization. The decrease in memory utilization observed across models with an increasing number of GPUs is in line with our expectation that batch size scaling is still important for efficient distributed training ~\cite{Hu2019}. Something to note is that while BERT sees about a two-fold reduction in memory utilization (from about 40\% to 20\%) as we scale up training from 2 to 128 GPUs, SM utilization barely changes (and slightly increases). 

Interestingly, the GNNs exhibit two extremes of GPU utilization: though SchNet has the lowest memory and SM utilization rates,  DimeNet, in contrast, has the highest memory utilization and nearly matches the SM utilization rates of BERT and VGG16---both heavily over-parameterized networks in their own respective domains. These observations suggest careful consideration and selection of GNN models when seeking to maximize benefits from scaling training across multiple GPUs, with all else equal. 


\begin{figure}[t]
    \centering
    \includegraphics[scale=0.48]{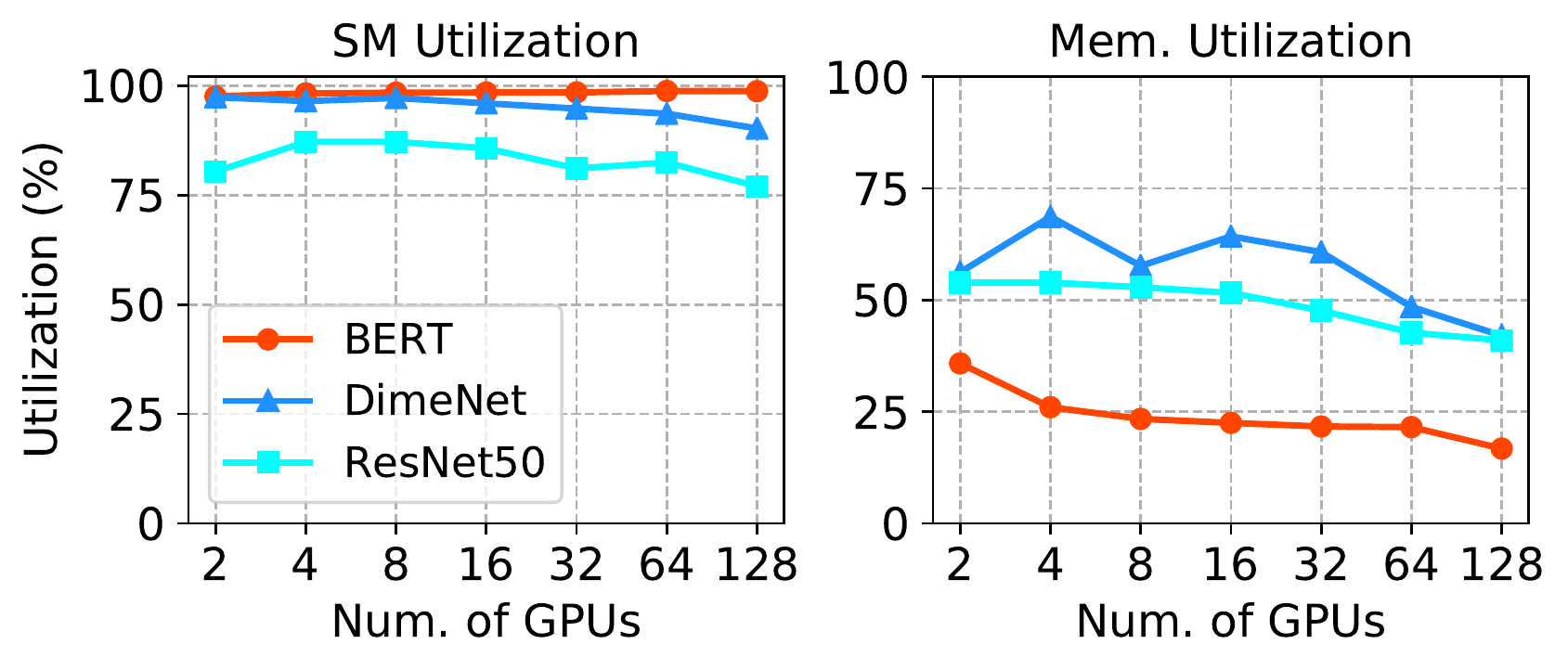}
    \caption{\textbf{GPU utilization decreases with a larger number of GPUs due to increasing communication overhead.} The effects of scaling distributed training on GPU SM and memory utilization for models trained on 2 - 128 GPUs. For simplicity, three representative models (BERT, DimeNet, and ResNet50) are shown. Other models (not shown here) show similar results.}
    \label{fig:char2}
\end{figure}

\begin{figure}[t]
    \centering
    \includegraphics[scale=0.48]{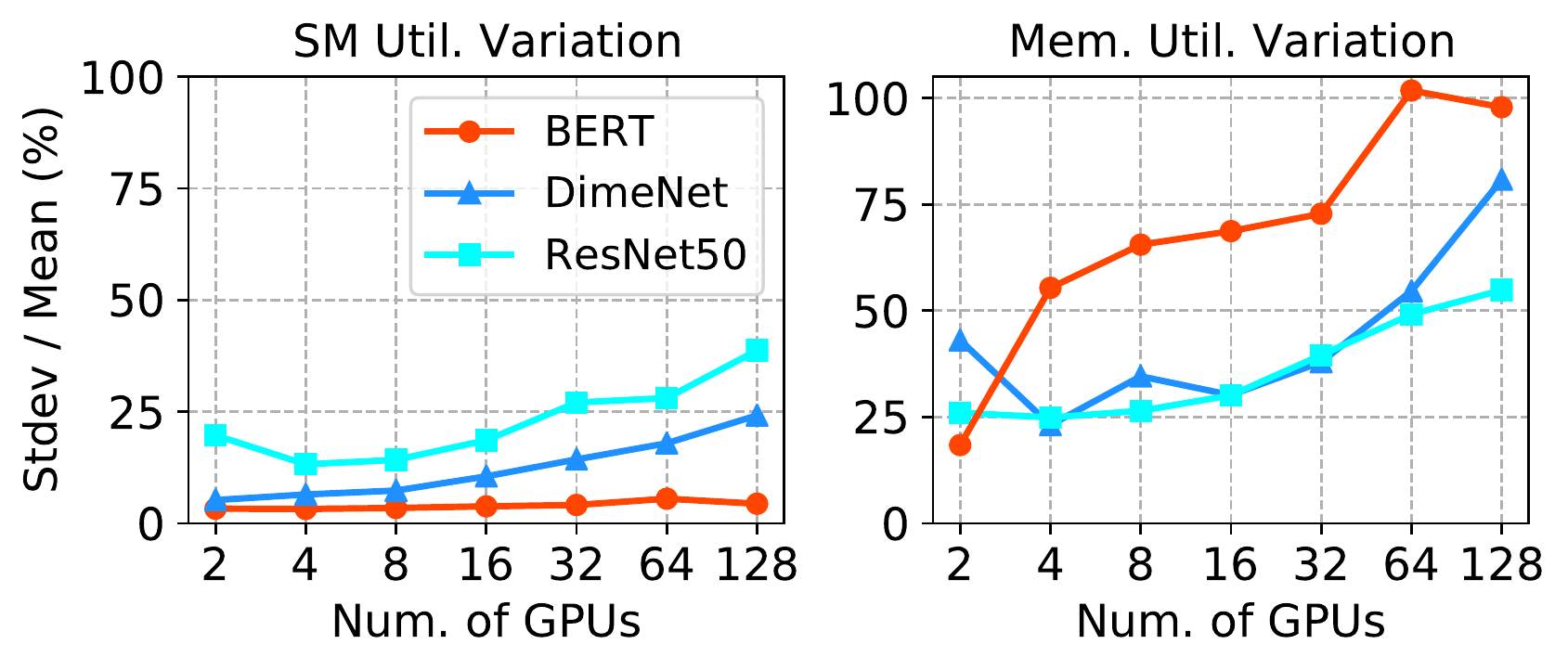}
    \caption{\textbf{Variation in GPU utilization during training increases with an increasing number of GPUs.} The coefficient of variation on GPU utilization rates during distributed training for BERT, DimeNet, and ResNet50 models across training for 2--128 GPUs. Other models (not shown here) show similar results.}
    \label{fig:char3}
\end{figure}

\noindent\newline\textbf{Though inefficiencies from distributed training become more pronounced with more GPUs due to higher communication overhead, they do so at different paces depending on the model; moreover, the variation in GPU utilization rates during training increases with more GPUs. }Figure ~\ref{fig:char2} shows the SM and memory utilization of three models as we scale up the number of GPUs used in distributed training from 2 to 128. The models shown tend to have high SM utilization (>75\%) but relatively lower memory utilization. As the number of GPUs increases, the memory utilization for all models decreases due to the increased communication overhead across GPU workers; this botteneck results in the memory controller having more idle time as it waits for the reduce step to complete. 

As such, we find that while memory utilization rates do decrease with an increased number of GPUs, the models do so at noticeably different rates (see right pane of Figure ~\ref{fig:char2}). While ResNet50 exhibits a continuous, gradual pace of decline in memory utilization rates, starting from about 50\% memory utilization as we scale up the number of GPUs, BERT sees a more significant decline in memory utilization even when scaling up from 2 GPUs to 4 GPUs before exhibiting slower declines until another more pronouced decline when scaling up from 64 to 128 GPUs. In contrast, DimeNet shows more varied, less monotonic behavior: rather than a continued decline in memory utilization rates with increased GPU counts, we see only continuous decrease in DimeNet's memory utilization beyond 16 GPUs. Even so, DimeNet shows the highest rate of memory utilization across all GPU counts, achieving memory utilization at least as high as, if not higher than, ResNet50, the model with the next highest but better behaved memory utilization rates across GPU counts. In terms of utilization trends that seem invariant to the number of GPUs, we note that although BERT has the lowest rate of memory utilization across all GPU counts, it has a near constant SM utilization at almost 100\% across all GPU counts.   

Since each training job has different phases during the processing of a mini-batch (e.g., forward propagation, backward propagation, aggregation), GPU utilization will naturally vary throughout these different phases of a single training run. To characterize this variation within or during the training process of each model, we calculate the coefficient of variation (CV) for GPU memory and SM utilization---defined as $CV = \sigma / \mu$ where the standard deviation of the utilization data ($\sigma$) is normalized by its mean ($\mu$) utilization---for each model's training lifetime as we scale up the number of GPUs. As we see in Figure ~\ref{fig:char3}, for all models, the variation in memory utilization as measured by CV increases with additional GPUs. We note that despite BERT's SM utilization rate itself being largely invariant to the number of GPUs used in training (left pane of Figure ~\ref{fig:char2}), the variation in its SM utilization rate during training is significantly higher with a $CV=100\%$ when scaled to beyond 32 GPUs (right pane of Fig.~\ref{fig:char3}). Overall, we see an increase in variation, as measured via CV, in both memory and SM utilization rates during training as the number of GPUs increases. Similar GPU utilization variance trends are seen with 200 W (Appendix \ref{appendix:figs} Figure \ref{fig:char3_200}) and 100 W (Appendix \ref{appendix:figs} Figure \ref{fig:char3_100}) power caps.

\begin{figure}[t]
    \centering
    \includegraphics[scale=0.45]{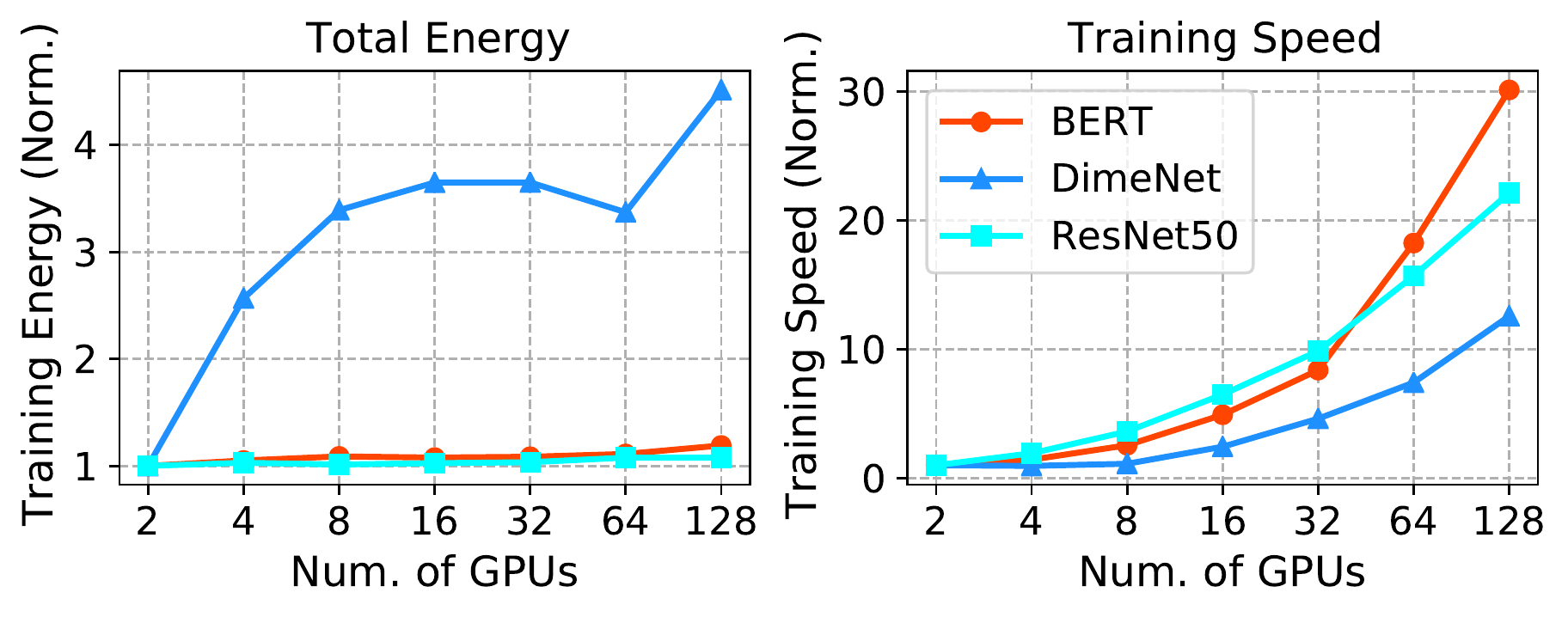}
    \caption{\textbf{Total energy consumption and training speed increase with increasing number of GPUs.} Total training energy and training speed are normalized by their counterparts under 2-GPU training.}
    \label{fig:char4}
\end{figure}

\noindent\newline\textbf{Trade-offs between total energy consumed during training and distributed training speed are model-dependent. }The goal of distributed training is to train a model faster; however,  more GPUs increase energy consumption, which is particularly concerning given the implications for its carbon footprint. Training on 128 V100 GPUs emits 22 kg of carbon dioxide an hour~\cite{lacoste2019quantifying}. 

Figure~\ref{fig:char4} shows the total energy consumption and training speed normalized by the energy and speed of 2-GPU training. The DimeNet \footnote{We note that the authors of DimeNet have released DimeNet++ \cite{klicpera2020fast}, which is an optimized architecture with faster training that may alleviate some scaling issues. However, DimeNet++ is not yet available in PyTorch Geometric and so was not considered in this study.} GNN shows poor scaling relative to the BERT NLP model and the ResNet CNN model. We observe similar behavior for the SchNet GNN (not shown in Figure~\ref{fig:char4} for brevity). The total energy consumption of DimeNet training quickly escalates to more than three times when increasing the number of worker GPUs, but this energy cost comes with the advantage of a $60\times$ speedup. On the other hand, both the BERT and ResNet models are still energy-friendly when trained at larger scales, incurring negligable amounts of additional energy required but still achieving more than a $30\times$ and $20\times$ speedup, respectively.

\begin{figure}[t]
    \centering
    \includegraphics[scale=0.44]{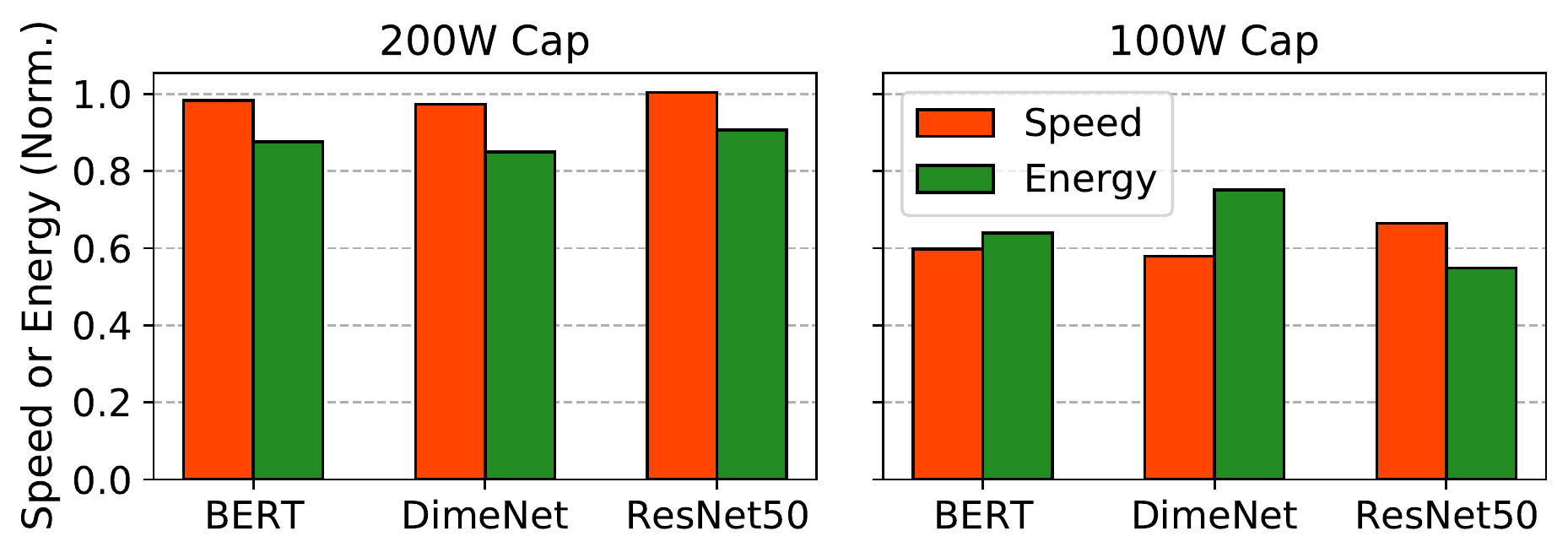}
    \caption{\textbf{Restricting GPU power to 200 W decreases total energy expenditure without impacting training time, while further restricting power to 100 W significantly degrades training time.} Training speed and total training energy on different GPU power capping settings when trained with 2 GPUs. Values are normalized to training speed and energy without power capping (i.e., 250 W).}
    \label{fig:char5}
\end{figure}

\noindent\newline\textbf{At optimal power capping settings, training speed suffers minimal impact but energy savings can be significant.} The V100 GPUs offer a wide range of power capping capabilities. In Figure~\ref{fig:char5}, we show the effects of power capping all GPUs at different levels (200 W and 100 W, the non-capped maximum power is 250 W). Training speed and energy in Figure ~\ref{fig:char5} are normalized by their non-capped (i.e., 250 W) counterparts. 

With capped power settings, slower training speeds are expected due to a lower resulting clock frequency. However, as total energy consumption depends on both the power and the total training time spent, our experiments aim to tease out the net effects on training time and identify potential opportunities for non-disruptive energy savings. In Figure~\ref{fig:char5}, we see that total energy consumption from each model's training is indeed reduced under power capping relative to the non-capped 250 W setting. We find an optimal power cap of 200 W where training speed degradation is negligible, but the energy savings are significant---at least 10\% across all three models, offering an optimal trade-off between training speed and energy savings for all models considered here. Capping the power further to 100 W results in further energy savings but training slows significantly, offering a less desirable energy-training tradeoff. Our experiments also reveal that by scaling up training with more GPUs in a distributed training context, the slowdown due to power capping becomes less significant (Appendix \ref{appendix:figs} Figure \ref{fig:char5_128gpu}), which may be the result of larger-scale training under-utilizing each GPU compared to smaller scale training.

Overall, we find that power limiting the GPUs offers an effective way of limiting energy consumption across a variety of deep learning models with minimal adverse impact on the time spent training. This effect was consistent across all models tested in this paper. Again, we note that our work does not examine the effect of this approach on test-set performance given a time and energy budget but this is the focus of ongoing and future work. From the operational aspects of a supercomputing center, our findings suggest a simple strategy for reducing energy consumption in a datacenter with minimal impact on computational burden. However, the availability of a mathematical model that can describe, and potentially predict, these effects in order to act as a guide for anticipatory scaling strategies would be an important tool in helping researchers and practitioners alike  in understanding these trade-offs. In the next section, we provide a preliminary effort to do so---we present a simple, descriptive model that expresses a statistically significant relationship underlying these trade-offs when scaling up distributed training across a different number of GPUs.